\documentclass[10pt,twocolumn,letterpaper]{article}

\usepackage{iccv}
\usepackage{times}
\usepackage{epsfig}
\usepackage{graphicx}
\usepackage{amsmath}
\usepackage{amssymb}
\usepackage{booktabs}
\usepackage{multirow}
\usepackage{placeins}
\usepackage{makecell}
\usepackage{algorithm}
\usepackage{algpseudocode}
\usepackage{authblk}
\usepackage{longtable}

\usepackage[pagebackref=true,breaklinks=true,letterpaper=true,colorlinks,bookmarks=false]{hyperref}
\usepackage[capitalize]{cleveref}
\crefname{section}{Sec.}{Secs.}
\Crefname{section}{Section}{Sections}
\Crefname{table}{Table}{Tables}
\crefname{table}{Tab.}{Tabs.}

\iccvfinalcopy 

\def\iccvPaperID{13} 

\ificcvfinal\fi
\begin{document}
\title{CoroNetGAN: Controlled Pruning of GANs via Hypernetworks}
\makeatletter \renewcommand\AB@affilsepx{\hfill \protect\Affilfont} \makeatother
\author[1]{Aman Kumar}
\author[1]{Khushboo Anand}
\author[1]{Shubham Mandloi}
\author[1]{Ashutosh Mishra}
\author[1]{Avinash Thakur}
\author[1]{Neeraj Kasera}
\author[2]{Prathosh A P}
\affil[1]{OPPO Mobiles R \& D Center, Hyderabad, India \thanks{E-mail:\{aman.kumar1, khushboo.anand, shubham.mandloi, ashutosh.mishra1\\avinash.thakur, neeraj.kasera\}@oppo.com}}
\affil[2]{Indian Institute of Science, Bangalore\thanks{E-mail:\{prathoshap@gmail.com\}}}
\maketitle
\ificcvfinal\thispagestyle{empty}\fi

\begin{abstract}
  Generative Adversarial Networks (GANs) have proven to exhibit remarkable performance and are widely used across many generative computer vision applications. However, the unprecedented demand for the deployment of GANs on resource-constrained edge devices still poses a challenge due to huge number of parameters involved in the generation process. This has led to focused attention on the area of compressing GANs. Most of the existing works use knowledge distillation with the overhead of teacher dependency.
  Moreover, there is no ability to control the degree of compression in these methods. Hence, we propose CoroNetGAN for compressing GAN using the combined strength of differentiable pruning method via hypernetworks. The proposed method provides the advantage of performing controllable compression while training along with reducing training time by a substantial factor. Experiments have been done on various conditional GAN architectures (Pix2Pix and CycleGAN) to signify the effectiveness of our approach on multiple benchmark datasets such as $\text{Edges}\rightarrow\text{Shoes}$,  $\text{Horse}\leftrightarrow\text{Zebra}$ and $\text{Summer}\rightarrow\text{Winter}$. The results obtained illustrate that our approach succeeds to outperform the baselines on $\text{Zebra}\rightarrow\text{Horse}$  and $\text{Summer}\rightarrow\text{Winter}$ achieving the best FID score of 32.3 and 72.3 respectively, yielding high-fidelity images across all the datasets. 
  Additionally, our approach also outperforms the state-of-the-art methods in achieving better inference time on various smart-phone chipsets and data-types making it a feasible solution for deployment on edge devices. 
\end{abstract}   

\section{Introduction}
\label{sec:intro}


Computer vision applications such as image-to-image translation, image synthesis, image generation, super resolution etc. have seen tremendous progress yielding high-fidelity images with the advent of GANs\cite{Goodfellow2014GenerativeAN}. The development of image-based GAN applications have in-turn accelerated the demand for deployment of such models on edge devices for the usage of the end consumers. However, the complexity of training such parameter heavy models to generate visually pleasing images result in high computational and memory overhead which acts as a bottleneck in deployment of GANs on mobile devices. For instance, the popular CycleGAN requires over 56.8G MACs (Multiply-Accumulate Operations) for generating a single image of resolution $256\times256$ pixels. On the other hand, Pix2Pix requires 18.6G MACs which is 4X compared to traditional Res-Net50~\cite{he2016deep} architecture. This huge number of operations is not desirable for the deployment on edge devices. Hence, there is a need for compressing these networks by removing the redundant parameters and reducing the memory and computational consumption. \par
Discriminative approaches such as image classification, object detection and semantic segmentation have been at the receiving end of undivided focus since these networks have surpassed human imagination but still, for the learning to saturate, these networks take huge amount of training time. For instance, the popular image classification model, Alexnet~\cite{Krizhevsky2012ImageNetCW} has 60 million parameters and requires about 240 MB of memory while VGG16~\cite{Simonyan2015VeryDC} has 130 million parameters and has takes around 500 MB of memory. The research community has given  unmitigated attention on the application of model compression techniques to accelerate deployment of image classification and object detection networks using techniques like weight quantization~\cite{Jacob2018QuantizationAT,Li2017PruningFF}, pruning~\cite{xie2020localization,zhang2018systematic} and
knowledge distillation~\cite{romero2014fitnets,chen2017learning}. However, these methods are not directly applicable to generative models such as GANs. A lot of the recently proposed methods  have tried to compress generative adversarial networks using the combined techniques of knowledge distillation~\cite{Aguinaldo2019CompressingGU,Chen2020DistillingPG} and channel pruning\cite{Wang2020GANSA, Liu2021ContentAwareGC}. However, these approaches don't allow controllable compression to happen neither using a single technique nor through the combination of multiple techniques. 


To address the above-mentioned issues, we propose a novel method for compressing the GAN using differentiable pruning method using the concept of hypernetwork. The compression is performed during the training regime. The proposed hypernetwork takes latent vector as an input and dynamically produces weights of a given layer of the generator network. This input latent vector decides the pruning rate for different layers in the network. Sparsification of latent vector is achieved via proximal gradient. Post sparsification, the latent vectors are passed through the hypernetwork that in turn generates the weight of the generator network. Since the latent vector and the weights of the generator network are covariant with each other, the sparsification of latent vectors leads to the pruning of the weights of the concerned network. The proposed method also helps in reducing the training time and inference time as compared to that of conventional GAN training method. Through the experiments on different conditional generative models on various datasets, the potential of the proposed method is revealed. The main contributions of the paper can be summarized as follows:
\begin{enumerate}
\item We propose CoroNetGAN, an approach based on differentiable pruning via hypernetworks for GAN compression. To the best of our knowledge, this is the first work that achieves model compression using controllable pruning via hypernetwork for conditional GANs. Our proposed approach compresses the GAN network in a controlled way by providing the compression rate as an input to the algorithm.  
\item Compression is achieved simultaneously alongside training unlike the distillation based methods that involve teacher dependency~\cite{Ren2021OnlineMD}. CoroNetGAN outperforms state-of-the-art compression technique~\cite{Ren2021OnlineMD} on training time on all the datasets validating the effectiveness of our technique both on training latency and visual appearance of the generated images. This will be of great advantage in reducing the training time while maintaining the accuracy when training GANs on bigger datasets containing billion of images. 
\item Our proposed approach, CoroNetGAN outperforms state-of-the-art conditional GAN compression methods on widely used $\text{Zebra}\rightarrow\text{Horse}$ and $\text{Summer}\rightarrow\text{Winter}$ datasets. CoroNetGAN obtains reasonable qualitative and quantitative results on other datasets. CoroNetGAN also outperforms state-of-the-art compression techniques~\cite{Ren2021OnlineMD} on inference time.
\end{enumerate}

\section{Related Work}
\subsection{Generative Adversarial Networks}

GANs~\cite{Goodfellow2014GenerativeAN} have proven to generate realistic results on a variety of tasks. For instance, Isola et al.~\cite{Isola2017ImagetoImageTW} propose Pix2Pix for paired image-to-image translation trained via the combination of adversarial loss and pixel-wise regression loss in order to ensure the visual quality of generated images. Later, ~\cite{wang2018high} is proposed that helps to increase the resolution of translated images with multi-scale neural networks and edge maps. GANs have also been proposed to perform image deblurring~\cite{kupyn2018deblurgan}, style transfer~\cite{li2021image, Liu_2021_CVPR}, image super resolution~\cite{wang2018esrgan} along with text-to-image generation~\cite{Tao18attngan}. Zhu et al.~\cite{Zhu2017UnpairedIT} propose CycleGAN for unpaired image-to-image translation. The algorithm trains generators on different domains of data through a weakly supervised setting using cycle consistency loss. The final objective is to convert the data from one domain to other without using any label information. 
\subsection{GAN Compression}
The tremendous resource consumption by GANs has garnered recent attention towards GAN compression. Wang et al.\cite{Wang2019QGANQG} proposes a novel quantization method and multi-precision quantization algorithm considering different sensitivities of discriminator and generator. Aguinaldo et al. \cite{Aguinaldo2019CompressingGU} introduces the idea of knowledge distillation in GANs between large over-parameterized network and small few parameter networks optimized using joint and mean squared error loss functions. However, the only focus here is to compress the generator keeping the discriminator intact. Most usage of GANs in mobile devices is based on the application of image-to-image translation task. \cite{Chen2020DistillingPG} distills the student discriminator to assist training of the student generator and also focused on image translation problem using Pix2Pix framework. Chang et al. in \cite{Chang2020TinyGANDB} focuses to mimic the functionality of BigGAN with a smaller compressed network and fewer parameters. Different devices with varied computing power require generators of different sizes. In order to accommodate this trade-off, Slimmable-GAN\cite{Hou2021SlimmableGA} proposes flexible switching between the multi-width configurations. Further, Ren et al.\cite{Ren2021OnlineMD} overcomes the complex multi-stage compression process and proposes a single-stage GAN online distillation strategy to obtain the compressed model. However, these approaches use images from the teacher directly to distill knowledge. Zhang et al.\cite{Zhang2022WaveletKD} proposes the idea of investigating GAN compression from frequency perspective and introduces the idea of wavelet analysis. They decompose the image into frequency bands and perform distillation only on bands with higher frequency unlike naive methods that do not prioritize the high frequency. \cite{Zhang2022RegionawareKD} aims to find crucial regions in the image using attention module. Considering the attention value important to the region, features are distilled from teacher to student. Recent works such as \cite{Jin2021TeachersDM} introduce an Inception-based Residual block replacing the original residual blocks in CycleGAN and search for student generator from teacher generator via pruning followed by Similarity based Knowledge Distillation. Further, approaches integrating various compression techniques are also proposed. \cite{Wang2020GANSA} combines model distillation, channel pruning and quantization and generate a unified optimization form which achieves superior trade-off compared with standalone compression techniques. Liu at al.\cite{Liu2021ContentAwareGC} combines the idea of channel pruning and knowledge distillation and mainly expands the focus on accelerating unconditional GANs.\par


\begin{figure*}
\centering
\includegraphics[width=0.95\textwidth]{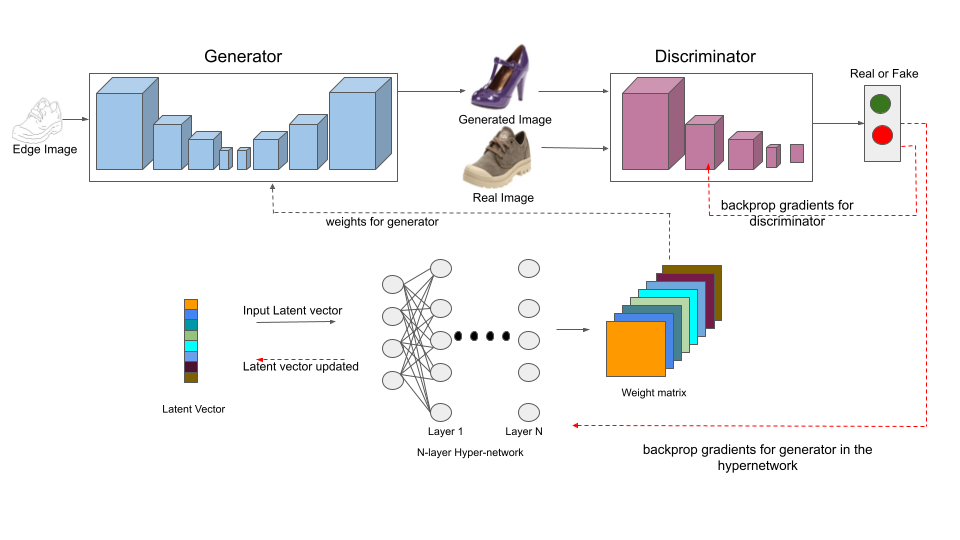}
\vspace*{-10mm}
\caption{Illustration of the proposed algorithm designed for compressing GAN's using controllable differentiable pruning. A latent vector is attached to each of the convolution layer of the generator. The latent vector generates the weights for the generator via hypernetwork. Sparsification of the latent vector leads to pruning of the corresponding weights of the generator network. The proposed design allows the latent vector and its corresponding weight matrix to be covariant with each other. The generator generates visual results using the computed weight matrix through the hypernetwork \textit{(Best viewed when zoomed)}.} \label{figure_block_diagram}
\vspace{-4.5mm}
\end{figure*}

\subsection{HyperNetworks}

Hypernetworks are a group of smaller networks that generates the weights for a larger network. These smaller neural networks have been used historically for vision\cite{article8}, functional representation\cite{article10} and bayesian inference tasks\cite{article11}.\par
Albeit the word hypernetwork has been coined recently, the concept of using dynamic parameter generation has been used by researchers for a long time~\cite{article1}. Von der Malsberg et al.\cite{article2} indicates a possibility of dynamic modelling between a slow classical weight and a dynamic decaying connection. The technique to model short term memory by computing weight changes of another network was initiated by Schmidhuber et al.\cite{article3}. Parameter prediction through co-relation between different parameters of the neural network was extensively studied in \cite{article4}. A weight matrix is produced using a learnable lower dimensional matrix using a linear operation.\cite{article5} uses weight matrices as a factored representation and feed forward one-shot learners reducing the dimensionality of the hypernetwork.\cite{article6} proposes an approach to calculate the parameters for image transformation using a weight generating network. ~\cite{article7} proposes an approach for generating weights for visual question answering task. The parameter prediction network takes input the questions post which the network predicts weights of the main network. In addition, they also use hashing of parameters to reduce the size of the final matrix of parameters. The concept of dynamic filters has been used for image super-resolution~\cite{article8}. These filters are computed based on input using a similar concept to hypernetwork.



\section{Methodology}
GAN consist of a generator and a discriminator network employed in a min-max game. The proposed method allows compression of the generator network while training. Compression is achieved using differentiable meta pruning which is based on the idea of hypernetwork. Hypernetwork is responsible for generating the weights of the generator network for each of its layer. The input to the hypernetwork is a latent vector and the output is a weight matrix of the generator network.\par
During the forward pass, latent vector is given as an input to hypernetwork to generate the weights of the generator. During back-propagation, the gradient flows in the hypernetwork instead of the main network. 
It is designed in a way such that its output is covariant with the input latent vector. Proximal Gradient helps in pruning of output channels of the generator network by eliminating the redundant parameters automatically.
\subsection{HyperNetwork Design}
The hypernetwork consists of three layers. The latent layer takes as input the latent vectors and computes a latent matrix. The embedding layer projects the elements of the latent matrix to an embedding space. The final layer converts the embedded vectors to the final output. The design is taken from \cite{li2020dhp}. As an example, consider the generator to be an L-layer convolutional neural network. Each layer of the network has its own corresponding latent vector that is responsible for generating the weights of the corresponding layer. The size of the latent vector is equal to the number of output channels in that layer. For instance, consider an \emph{$l$}-th convolutional layer having \emph{$n*c*w*h$} number of parameters, where \emph{n} and \emph{c} are the output and input channels and \emph{w*h} is the size of kernel respectively.
Suppose that the latent vector corresponding to that particular \emph{$l-th$} layer is \emph{$v^l$} $\in$ \emph{$R^c$}. Therefore, the previous layer has latent vector \emph{$v^{l-1}$} $\in$ \emph{$R^n$}. The hypernetwork takes latent vector of current layer (\emph{$v^l$}) and its previous layer (\emph{$v^{l-1}$}) as input and will output the weights matrix of the \emph{l-th} layer of the generator network. Initially, the first layer of the hypernetwork computes a latent matrix using the two latent vectors:
\begin{equation}
\textbf{V}^l  =  \textbf{v}^l . \textbf{v}^{{l-1}^T}  +  \textbf{B}_0
\label{eq:eq1}
\end{equation}
where,
$$\textbf{V}^l,  \textbf{B}_0 \in R^{n*c}$$
Here, [\emph{T}] denotes transpose of the matrix while [.] denotes matrix multiplication.

Subsequently, the second layer of the hypernetwork projects every element of the latent matrix to a m-dimensional embedding space as follows:
\begin{equation}
\textbf{S}^{l}_{i j} = \textbf{V}^l_{i j}\textbf{w}^l_1 + \textbf{b}^1_l   \hspace{0.5cm}  i = 1..n, j = 1...c
\label{eq:eq2}
\end{equation}
\vspace{0.15cm}
where,
$$\textbf{S}^{l}_{i j}, \textbf{w}^l_1, \textbf{b}^1_l \in R^m$$

Here, we are considering $\textbf{w}^l_1$ and $\textbf{b}^1_l$ as different for different elements of the matrix. The subscript (\emph{i, j}) has not been used for easier interpretation of the above mentioned equations.
The vectors  $\textbf{w}^l_1$, $\textbf{b}^1_l$ and $\textbf{S}^{l}_{i j}$ for all the elements of the matrix together forms a 3D tensor, i.e., $\textbf{W}^l_1$, $\textbf{B}^l_1$ and $\textbf{S}^l_1$ $\in$ $R^{n*c*m}$.\\

After the second step, the final layer of the hypernetwork is responsible for converting the embedding vectors to the output($\textbf{F}^{l}_{i j}$) which can be used as weight matrix of the Generator network. This is done by multiplying the embedded vectors $\textbf{S}^{l}_{i j}$ by an explicit matrix as follows:
\begin{equation}
\textbf{F}^{l}_{i j} = \textbf{w}^l_2 . \textbf{S}^l_{i j}+ \textbf{b}^2_l   \hspace{0.5cm}  i = 1..n, j = 1...c
\label{eq:eq3}
\end{equation}


where,
   $$\textbf{F}^{l}_{i j}, \textbf{b}^l_2 \in R^{wh}$$ 
\hspace{0.5cm}and,
   $$\textbf{w}^l_2 \in R^{wh*m}$$
\\
 $\textbf{w}^l_2$ and $\textbf{b}^2_l$ are different and unique for each of the element and subscript (\emph{i,j}) has not been used for easier interpretations. Vectors $\textbf{w}^l_2$, $\textbf{b}^2_l$ and $\textbf{F}^{l}_{i j}$ for all the elements together will be high-dimensional tensors i.e., $\textbf{W}^l_2$ $\in$ $R^{n*c*wh*m}$ and $\textbf{B}^l_2$ and $\textbf{F}^l$ $\in$ $R^{n*c*wh}$.
 \\

Combining~\ref{eq:eq1},~\ref{eq:eq2} and~\ref{eq:eq3}, the functionality of the proposed approach can be collectively written as:
\begin{equation}
\textbf{F}^l = h(\textbf{v}^l, \textbf{v}^{l-1}; \textbf{W}^l, \textbf{B}^l),  
\end{equation}
where,
h(.) denotes the functionality of the above architecture. The final output $\textbf{F}^l$ will be used as the weight parameter of the \emph{l}-th layer. The hypernetwork is designed in such a way that the weight matrix of the generator is covariant with it's corresponding input latent vector as pruning an element in the latent vector automatically leads to removal of corresponding slice in  the final weight matrix( $\textbf{F}^l$). Figure~\ref{figure_block_diagram} depicts the overall workflow of the proposed CoroNetGAN.\par


While designing the hypernetwork, we also execute residual connections in the network. In case of residual or skip connections, we take the input latent vector as the combination of the latent vector of the previous layer and the corresponding layer from which the skip connection originates. We concatenate both the input latent vectors to create one single input latent vector. The resultant input latent vector along with latent vector of the current layer is used to create the latent matrix. The creation of latent matrix is followed by execution of steps((2),(3)) for generating the weights matrix of the convolution layer.
\subsection{Vector Sparsity using Proximal Gradient}
The differentiable property of the algorithm comes through the use of proximal gradient. Proximal gradient helps in sparsification of the latent vector by searching the potential candidates. Since latent vector is covariant with the weight matrix of the Generator network, it leads to compression of the Generator network. During training time, the parameters of the hypernetwork is updated using Stochastic Gradient Descent (SGD) optimization algorithm. During back-propagation, gradients flow from the Generator network to the hypernetwork. The latent vectors are updated using the proximal gradient \cite{li2020dhp} which leads to sparsified input latent vectors as follows:
\begin{equation}
\textbf{v}[k+1] = \textbf{prox}_{\lambda \mu R}(v[k] - \lambda \mu \nabla \emph{L}(v[k]))    
\end{equation}


The proximal gradient algorithm forces the potential elements of the latent vectors to approach zero quicker than the others without any human effort and interference in this process. Due to the fact that proximal operator has closed-form solution and use of SGD, the whole solution is recognized as approximately differentiable.


\subsection{Network Pruning}
\begin{figure*}
\centering
{\includegraphics[width=.23\textwidth,height=.15\textwidth]{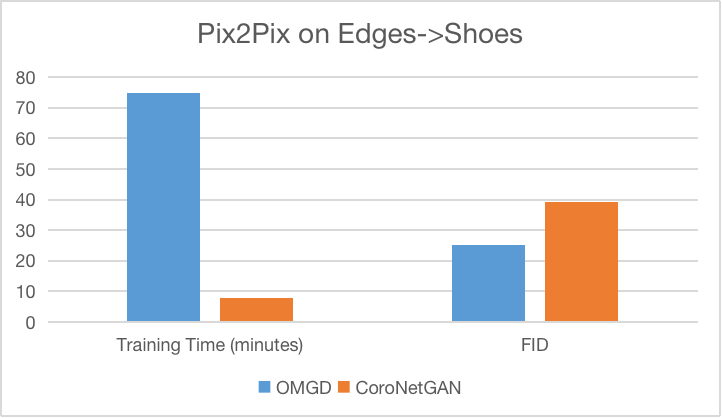}}
\hspace{0.3cm}
{\includegraphics[width=.26\textwidth]{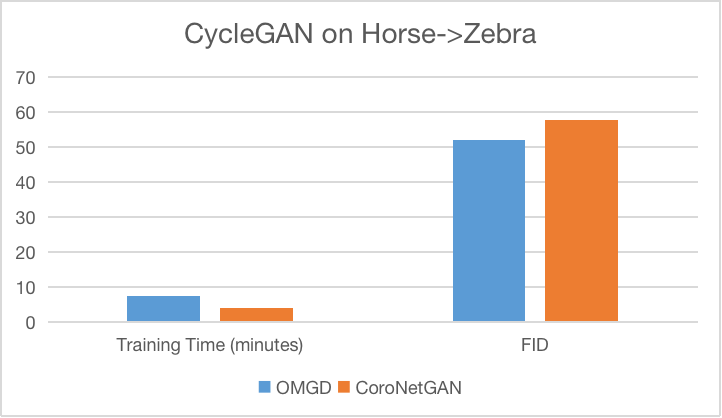} }%
\hspace{0.3cm}
{\includegraphics[width=.26\textwidth]{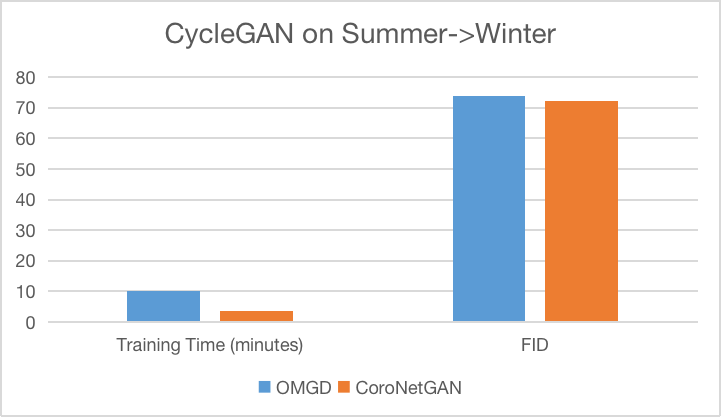} }%
\caption{Graphical representation of training time(in minutes) and FID for Pix2Pix(left) on $\text{Edges}\rightarrow\text{Shoes}$ and CycleGAN(middle,right) on $\text{Horse}\rightarrow\text{Zebra}$ and $\text{Summer}\rightarrow\text{Winter}$ datasets respectively. From the graphs, it is evident that total training time for our proposed approach is significantly lesser compared to OMGD~\cite{Ren2021OnlineMD}. For CycleGAN on $\text{Summer}\rightarrow\text{Winter}$ dataset, our algorithm outperforms OMGD~\cite{Ren2021OnlineMD} on both training time and FID \textit{(Best viewed when zoomed)}.}
\label{fig:graphs}
\end{figure*}
Our proposed method allows the weight matrix to be covariant with the it's corresponding latent vector. Hence, sparsifictaion of latent vector leads to the pruning of the corresponding weights of the CNN layer in the generator network. Our training regime consists of two stages, namely searching stage and converging stage. During the searching stage, proximal gradients helps in identifying the potential candidates of the latent vector. Therefore, after the searching stage, we get the sparsified latent vector (\emph{$\hat{v^l}$}). Proximal gradient help in elements of the latent vector either be zero or approaching towards zero. We use a mask(\emph{$m^l$}) on the sparsified latent vector with a predefined threshold($\tau$). This is followed by masking operation that compares every element of the latent vector with the threshold value. If greater than threshold, the returned value is one else zero. The sparsified latent vector, \emph{$\hat{v^l}$} is pruned with the help of the computed mask (\emph{$m^l$}).\par
Once the target compression ratio is achieved, the algorithm shifts from searching to the converging stage. In the converging stage, hypernetwork is discarded, and the training of the generator follows the conventional GAN training procedure. Upon extensive experimentation, it is observed that the number of epochs in searching stage is much smaller than the number of epochs in the converging stage. The pseudo-code of the proposed algorithm is mentioned in Algorithm~\ref{algorithm_main}.
\begin{algorithm}
\caption{CoroNetGAN Pseudo Code}\label{alg:cap}
\begin{algorithmic}
\State $total\_epochs \gets total\_number\_of\_epochs$
\State $targetflops \gets target\_compression\_ratio$
\State $latent\_vectors(v_1, v_2, .., v_i)$
\State $converging \gets False$
\State $epochs \gets 0$
\\
\State $\textbf{Compression via Differentiable Pruning}$
\While{$converging \neq True$}
\State $\bullet$ Sample \textbf{m}\{$z_1, z_2,..,z_i$\} images from given dataset
\State $\bullet$ Sample \textbf{m}\{$x_1, x_2,..,x_i$\} ground-truths
\State $\bullet$ Update Hypernetwork using SGD
\State\hspace{0.35cm} $\triangledown_{\theta_{h}} \dfrac{1}{m}\sum_{i=1}^{m}log(1-D(G(z_i)))$
\State $\bullet$ Update Discriminator using SGD
\State\hspace{0.35cm} $\triangledown_{\theta_{d}} \dfrac{1}{m}\sum_{i=1}^{m}\{logD(x^i)+ log(1-D(G(z_i)))\}$
\State $\bullet$ Compress latent vector using proximal gradient
\State \hspace{0.35cm} $\textbf{v}[k+1] = \textbf{prox}_{\lambda \mu R}(v[k] - \lambda \mu \nabla \emph{L}(v[k]))$
\State $\bullet$ $epochs \gets epochs+1$
\If {$flops-target\_flops \leq threshold$}
    \State $converging \gets True$
\EndIf
\If {$epochs \leq total\_epochs$}
    \State $break$
\EndIf
\EndWhile
\\
\State $\textbf{ Finetuning}$
\While{$epochs \leq total\_epochs$}
\State $\bullet$ Sample \textbf{m}\{$z_1, z_2,..,z_i$\} images from given dataset
\State $\bullet$ Sample \textbf{m}\{$x_1, x_2,..,x_i$\} ground-truths
\State $\bullet$ Update Generator using SGD
\State\hspace{0.35cm} $\triangledown_{\theta_{g}} \dfrac{1}{m}\sum_{i=1}^{m}log(1-D(G(z_i)))$
\State $\bullet$ Update Discriminator using SGD
\State\hspace{0.35cm} $\triangledown_{\theta_{d}} \dfrac{1}{m}\sum_{i=1}^{m}\{logD(x^i) + log(1-D(G(z_i)))\}$
\State $\bullet$ $epochs \gets epochs+1$
\EndWhile
\end{algorithmic}
\label{algorithm_main}
\end{algorithm}

\section{Experiments}
\subsection{Experiment Setting}
\subsubsection{Models and Datasets}
We evaluate our approach incorporating the following models to demonstrate the effectiveness of the proposed method:
\begin{enumerate}
  \item Pix2Pix~\cite{Isola2017ImagetoImageTW} for paired image-to-image translation with original U-Net generator architecture.
  \item CycleGAN~\cite{Zhu2017UnpairedIT} for unpaired image-to-image translation using Res-Net architecture to perform transformation on an image belonging to source domain to desired target domain.
  \item Deep Convolutional Generative Adversarial Network (DCGAN)~\cite{Radford2016UnsupervisedRL} that uses convolutional and convolutional-transpose layers in the discriminator and generator, respectively. 
\end{enumerate}
For the purpose of quantitative and qualitative evaluation, four datasets are utilised including $\text{Edges}\rightarrow\text{Shoes}$, $\text{Horse}\leftrightarrow\text{Zebra}$,  $\text{Summer}\rightarrow\text{Winter}$ and CIFAR10.
\begin{enumerate}
    \item $\text{Edges}\rightarrow\text{Shoes}$~\cite{Isola2017ImagetoImageTW}  is a paired image-to-image translation dataset including images edges of shoes to be mapped to their corresponding complete image of shoes. The dataset consists of 49825 images.
    \item $\text{Horse}\leftrightarrow\text{Zebra}$~\cite{Zhu2017UnpairedIT}   dataset contains images originally from ImageNet~\cite{deng2009imagenet}. It is an unpaired image-to-image translation dataset used for translating horse images to zebra and vice versa. In our experiments, the training set includes 1067 horse images and 1334 zebra images.
    \item $\text{Summer}\rightarrow\text{Winter}$~\cite{Zhu2017UnpairedIT}   is also unpaired image-to-image translation dataset which translates summer images to winter. We have used 1231 summer images for training purpose.
    \item CIFAR10 dataset~\cite{krizhevsky2009learning} consists of 50000 training images and 10000 test images across 10 different classes.

\end{enumerate}

\noindent Our approach CoroNetGAN with Pix2Pix architecture is bench-marked on  $\text{Edges}\rightarrow\text{Shoes}$  dataset. On the other hand, CoroNetGAN with CycleGAN has been bench-marked on $\text{Horse}\leftrightarrow\text{Zebra}$ and $\text{Summer}\rightarrow\text{Winter}$ datasets. \par
Although our proposed approach focuses on the compression for conditional GAN, we also made initial attempts to perform compression using our proposed algorithm for unconditional GAN (specifically DCGAN). 

\begin{figure}
    \centering
    \includegraphics[width=.11\textwidth]{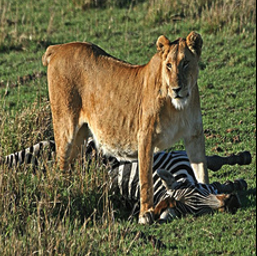}
    \includegraphics[width=.11\textwidth]{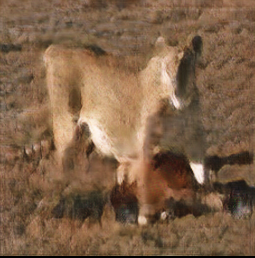}
    \includegraphics[width=.11\textwidth]{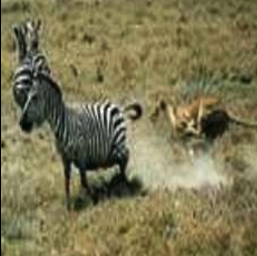}
    \includegraphics[width=.11\textwidth]{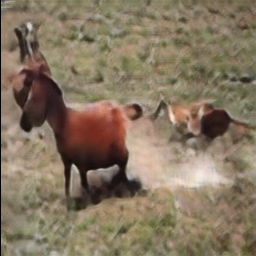}\\
    \includegraphics[width=.11\textwidth]{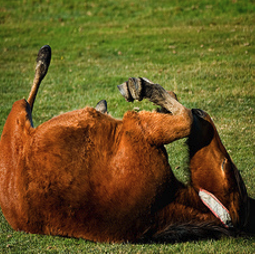}
    \includegraphics[width=.11\textwidth]{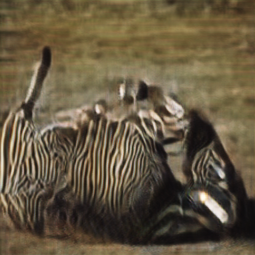}
    \includegraphics[width=.11\textwidth]{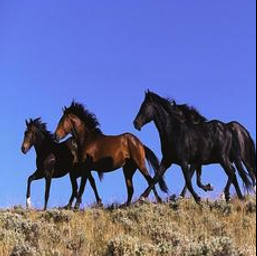}
    \includegraphics[width=.11\textwidth]{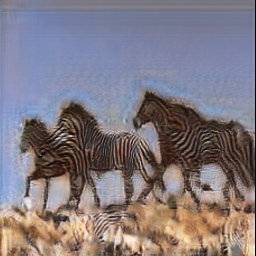}
    \caption{Samples generated from our approach. First row contains translated images from $\text{Zebra}\rightarrow\text{Horse}$ dataset. The second row contains translated images from $\text{Horse}\rightarrow\text{Zebra}$ dataset \textit{(Best viewed when zoomed).}}
    \label{fig:sample}
    \vspace{-1em}
\end{figure}

\begin{figure}
\includegraphics[width=\linewidth,height=0.55\linewidth]{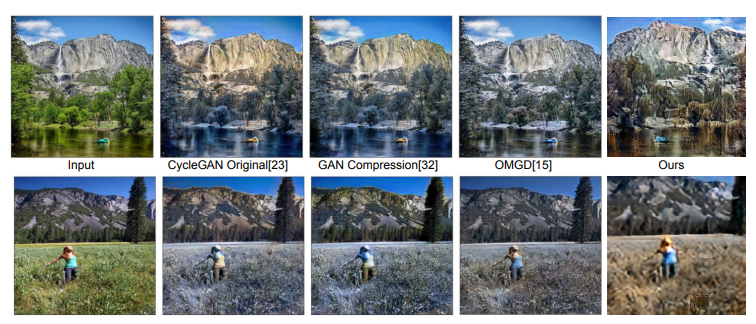}
\caption{Qualitative comparison of CoroNetGAN with CycleGAN architecture on  $\text{Summer}\rightarrow\text{Winter}$
dataset compared with original CycleGAN~\cite{Zhu2017UnpairedIT}, GAN Compression~\cite{Li2020GANCE} and OMGD~\cite{Ren2021OnlineMD} algorithms. Our approach generates visually realistic images and outperforms all the other algorithms on the FID metric \textit{(Best viewed when zoomed)}.} \label{cycle_gan_res}
\vspace{-4mm}
\end{figure}

\subsubsection{Implementation Details}
We train our proposed approach using single NVIDIA Tesla V100 GPU on PyTorch deep learning framework. For the algorithm to compress the network, a target compression ratio needs to be selected. When the difference between the actual compression and the target compression ratio falls below 2$\%$, pruning stops and model moves to fine-tuning state from the compression state. The number of parameters in the hypernetwork is proportional to the size of the embedding space. For the experimentation, embedding space is set to 8 across all the experiments. Learning rate is set to 0.0002. Batch size has been set to 4 and 1 for Pix2Pix and CycleGAN respectively on all the experiments across different datasets. The  sparsity regularization factor for the proximal gradient is set to 0.5 across all the experiments.

\subsubsection{Evaluation Setting}
For the quantitative performance comparison, we adopt Frechet Inception Distance (FID)~\cite{heusel2017gans} as common evaluation metric. FID is specifically developed for assessing the performance of GANs. It is used to evaluate the quality of the images generated by generative models by comparing the distribution of features corresponding to real and generated images using an InceptionV3~\cite{szegedy2016rethinking} network. A lower FID score is an indicator of high similarity between both the distributions and thus better quality of generated images. We have evaluated FID for different architectures with our approach on multiple datasets and compared it against existing methodologies.


\begin{figure}
\includegraphics[width=0.5\textwidth,height=0.2\textwidth]{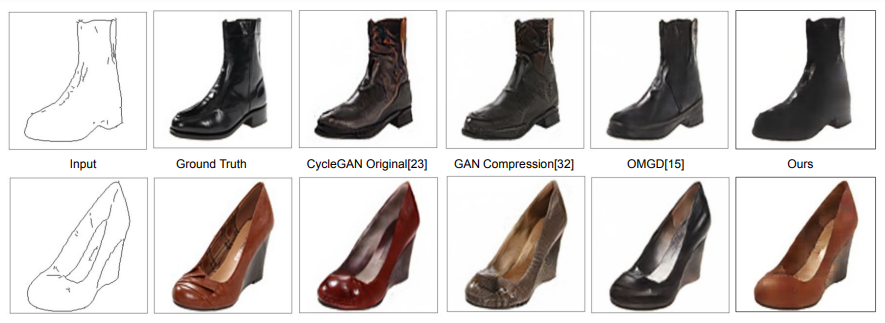}
\caption{Qualitative comparison of CoroNetGAN with Pix2Pix architecture on $\text{Edges}\rightarrow\text{Shoes}$
dataset compared with original Pix2Pix~\cite{Isola2017ImagetoImageTW}, GAN Compression~\cite{Li2020GANCE} and OMGD~\cite{Ren2021OnlineMD} algorithms. Our approach generates visually plausible images compared to state-of-the-art methods \textit{(Best viewed when zoomed in)}.} \label{pix2pix_gan_res}
\end{figure}

\begin{table*}[t!]
\begin{center}
\resizebox{0.8\textwidth}{!}{

\begin{tabular}{l l l l l l l }
    \hline
     \textbf{Model}   &  \textbf{Dataset} &  \textbf{Paper} &  \textbf{Params(M)}  & \textbf{FLOPs(G)} & \textbf{MACs(G)} & \textbf{FID} \\  
    \hline
    
\multirow{7}{*}{Pix2Pix} & \multirow{7}{*}{$\text{Edges}\rightarrow\text{Shoes}$} &

Original~\cite{Isola2017ImagetoImageTW}  & 54.4 & \textendash & 18.6 & 34.31 \\ \cline{3-7} 


&  & Region-Aware~\cite{Zhang2022RegionawareKD}     & 13.61 (4.00×) & 1.56 & \textendash &  77.69±3.14  \\

 &    &  Wavelet KD ~\cite{Zhang2022WaveletKD}   & 13.61 (4.00×) & 1.56 & \textendash &  80.13±2.18 \\ 
 
 &    &  DMAD ~\cite{Li2022LearningEG}   & 2.13 (25.5×)  & \textendash & 2.99 (6.2×) &  46.95\\ 
 
  &    & OMGD ~\cite{Ren2021OnlineMD} & 3.404 (16.0×) & \textendash &  1.219 (15.3×) & \textbf{25}\\ 
  
 &    &    \textbf{CoroNetGAN(75\%)}   & 13.225  & 4.8879   & \textendash  &  39.1\\ 
 &    &   \textbf{CoroNetGAN(95\%)}    & 4.721 & 1.2551  & \textendash  &  54.3\\ \cline{1-7}

\multirow{19}{*}{CycleGAN} & \multirow{9}{*}{$\text{Horse}\rightarrow\text{Zebra}$} &

Original~\cite{Zhu2017UnpairedIT}   & 11.3 & \textendash & 56.8  &  61.53    \\ \cline{3-7}

&  &  Region-Aware~\cite{Zhang2022RegionawareKD}    & 1.61 (7.08×) & 7.29 & \textendash &  60.01±5.22  \\

 &    &    Wavelet KD ~\cite{Zhang2022WaveletKD} & 1.61 (7.08×) & 7.29 & \textendash &  61.65±4.73 \\ 
 
&    &  DMAD ~\cite{Li2022LearningEG}   & 0.42 (26.9×)  & \textendash & 3.97 (14.3×) &  62.41\\ 

&    &  Teachers Do More Than Teach ~\cite{Jin2021TeachersDM}   & \textendash  & \textendash & 2.56 &  53.48\\ 

&    &  GAN Compression ~\cite{Li2020GANCE}   & 0.34 (33.3×)  & \textendash & 2.67 (21.2×) &  64.95\\ 

&    & Revisiting Discriminator in GAN Compression ~\cite{Li2021RevisitingDI}    & \textendash  & \textendash & 2.40 &  59.31\\ 

 &    & OMGD ~\cite{Ren2021OnlineMD}  & 0.137 (82.5×)   & \textendash & 1.408 (40.3×) & \textbf{51.92}\\ 
 
 &    &     \textbf{CoroNetGAN(75\%)}  & 2.685  & 0.217  & \textendash  &  57.7\\ 
 
  &    &     \textbf{CoroNetGAN(85\%)}  & 1.670  & 0.1347  & \textendash  &  60.9\\ \cline{2-7}

& \multirow{5}{*}{$\text{Zebra}\rightarrow\text{Horse}$} &

Original~\cite{Zhu2017UnpairedIT}  &  11.3 & 49.64 & \textendash &  138.07±4.01   \\\cline{3-7} 

&  &  Region-Aware~\cite{Zhang2022RegionawareKD}      & 1.61 (7.08×) & 7.29 (6.80×) & \textendash &  137.03±3.03  \\

&    &    Wavelet KD ~\cite{Zhang2022WaveletKD}  & 1.61 (7.08×) & 7.29 (6.80×) & \textendash &  138.84±1.47 \\ 

&    &  DMAD ~\cite{Li2022LearningEG}   & 0.30 (37.7×)  & \textendash & 3.50 &  139.3\\ 

  &   & \textbf{CoroNetGAN (75\%) }  & 2.685  & 0.217 & \textendash & \textbf{ 32.3} \\\cline{2-7}
  
& \multirow{5}{*}{$\text{Summer}\rightarrow\text{Winter}$} &

Original~\cite{Zhu2017UnpairedIT}    & 11.3  & \textendash & 56.8  &  79.12  \\\cline{3-7}

&    &   DMAD ~\cite{Li2022LearningEG} & 0.24 (47.1×)  & \textendash & 3.18 (17.9×) &  78.24\\ 

 &    & OMGD ~\cite{Ren2021OnlineMD}    & 0.137 (82.5×)   & \textendash & 1.408 (40.3×) & 73.79\\ 
 
 &    &  Auto-GAN ~\cite{Fu2020AutoGANDistillerST}     & \textendash & 4.34 & \textendash &  78.33\\ 
 
&   & \textbf{CoroNetGAN (75\%)}   & 2.685  & 0.217  & \textendash & \textbf{ 72.3} \\
&   & \textbf{CoroNetGAN (85\%)}   & 1.670 & 0.1347  & \textendash & { 74.7} \\\cline{2-7}

    \hline
    
\end{tabular}}

\end{center}

\caption{Performance comparison of CoroNetGAN with state-of-the-art algorithms on Pix2Pix and CycleGAN architectures. It can be observed that our approach achieves best FID with CycleGAN on $\text{Zebra}\rightarrow\text{Horse}$ and $\text{Summer}\rightarrow\text{Winter}$ datasets. Our results also outperform  \cite{Zhang2022RegionawareKD,Zhang2022WaveletKD,Li2022LearningEG,Aguinaldo2019CompressingGU,Li2021RevisitingDI} and achieve competitive FID on $\text{Horse}\rightarrow\text{Zebra}$ dataset. We also achieve second best FID score on $\text{Edges}\rightarrow\text{Shoes}$ dataset beating the results of \cite{Zhang2022RegionawareKD,Zhang2022WaveletKD,Li2022LearningEG}.}
\label{tab:multicolts1}
\end{table*}



    

\begin{table*}[t!]
\begin{center}
\resizebox{0.7\textwidth}{!}{

\begin{tabular}{l l l l l l}
    \hline
\textbf{Model} & \textbf{Dataset}  & \textbf{Method} & \textbf{Params(M)} & \textbf{FLOPs(G)}  & \textbf{FID} \\  
    \hline
   
\multirow{4}{*}{Pix2Pix} & \multirow{4}{*}{$\text{Edges}\rightarrow\text{Shoes}$} & Original~\cite{Isola2017ImagetoImageTW}   & 54.41 & \textendash  &  34.31  \\
\cline{3-6} 
& & \textbf{CoroNetGAN(75\%)} & 13.225 (24.31\%) & 4.8879 (26.94\%)  &  39.1\\ 
 
 \cline{3-6}
& & \multirow{2}{*}{\textbf{CoroNetGAN(G + D)(75\%)}} & Generator 13.321 (24.48\%) & Generator 4.8993 (27\%)  & \multirow{2}{*}{38.6}  \\

 & &  & Discriminator 0.725 (26.23\%) & Discriminator 0.4767 (26.74\%)  & \\ \cline{2-5} 

     \hline
\end{tabular}}

\end{center}
\caption{Generator and Discriminator compression in CoroNetGAN in Pix2Pix architecture on $\text{Edges}\rightarrow\text{Shoes}$ dataset. It is evident that compressing both generator and discriminator helps in improving the FID score.} 
\label{tab:multicolts2}
\end{table*}

\begin{table*}[t!]
\begin{center}
\resizebox{0.5\textwidth}{!}{
\begin{tabular}{l l l l}
    \hline
    \textbf{Chipset}  &  \textbf{d-type} & \textbf{Model} & \textbf{GPU Inference Time(CL)(ms)} \\
    \hline
\multirow{6}{*}{Qualcomm Snapdragon SM8450} & \multirow{2}{*}{32-bit} & \textbf{Ours} & {\hfil 12.5419} \\ 
&& OMGD~\cite{Ren2021OnlineMD} & \hfil15.3378\\
\cline{3-4}
& \multirow{2}{*}{16-bit} & \textbf{Ours}  & \hfil11.794 \\ 
&& OMGD~\cite{Ren2021OnlineMD} & \hfil15.283\\
\cline{3-4}
& \multirow{2}{*}{8-bit} & \textbf{Ours}  & \hfil12.244 \\ 
&& OMGD~\cite{Ren2021OnlineMD} & \hfil16.0191\\
\cline{1-4}
\multirow{6}{*}{Dimensity 1200-Max Octa} & \multirow{2}{*}{32-bit} & \textbf{Ours}  & \hfil20.7268 \\ 
&& OMGD~\cite{Ren2021OnlineMD} & \hfil21.1919\\
\cline{3-4}
& \multirow{2}{*}{16-bit} & \textbf{Ours}  & \hfil20.1961 \\ 
&& OMGD~\cite{Ren2021OnlineMD} & \hfil20.9635\\\cline{3-4}
& \multirow{2}{*}{8-bit} & \textbf{Ours}  & \hfil20.9972 \\ 
&& OMGD~\cite{Ren2021OnlineMD} & \hfil21.541\\
\cline{3-4}
\hline
\end{tabular}
}
\end{center}
\caption{Shows inference time comparison between the model compressed by our methodology to 95\% and the model compressed from \textbf{OMGD} on different processors. Both the models are trained on $\text{Edges}\rightarrow\text{Shoes}$ dataset. The inference time is computed for the input resolution of 256$\times$256. The 8-bit quantization of the compressed model results in increased processing time due to the presence of a higher number of quantization and de-quantization blocks compared to other data types.}
\label{tab:multicolts3}
\end{table*}

\subsection{Experimental Results}
\subsubsection{Quantitative Results}

We evaluate our approach on different models and datasets using evaluation setting mentioned in the previous section and report quantitative results compared with the corresponding state-of-the-art methods. The results can be summarized as follows:
\par
\textbf{Pix2Pix:}
We incorporate Pix2Pix with its original U-Net architecture in our proposed approach and report the experimental results in Table \ref{tab:multicolts1}. We observe that our approach is able to achieve second best FID score on $\text{Edges}\rightarrow\text{Shoes}$ dataset. We are also able to outperform the results of \cite{Zhang2022RegionawareKD,Zhang2022WaveletKD,Li2022LearningEG} by achieving a better FID. Although, our FID score is higher than \cite{Ren2021OnlineMD} but our approach outperforms it in terms of training time. Figure~\ref{fig:graphs} illustrates that our approach significantly improves training time however elevates FID score compared to~\cite{Ren2021OnlineMD}.

\par
\textbf{CycleGAN:}
Similar to previous works, we include Res-Net style CycleGAN in our method and report the results in Table \ref{tab:multicolts1}. We observe that the results on $\text{Zebra}\rightarrow\text{Horse}$ dataset outperform all the state-of-the-art approaches by achieving the best FID score of 32.3 corresponding to 75\% compression. Additionally, our approach is also able to improve over all the existing baselines by achieving FID score of 72.3 on $\text{Summer}\rightarrow\text{Winter}$ dataset. As illustrated in Figure \ref{fig:graphs} we also outperform \cite{Ren2021OnlineMD} on both training time and FID.
\par
Furthermore, we also observe that our approach with CycleGAN is able to beat the results of \cite{Zhang2022RegionawareKD,Zhang2022WaveletKD, Li2022LearningEG,Li2020GANCE,Li2021RevisitingDI} on $\text{Horse}\rightarrow\text{Zebra}$ dataset. Even though, we achieve greater FID score than \cite{Ren2021OnlineMD,Jin2021TeachersDM}, CoroNetGAN outperforms \cite{Ren2021OnlineMD} on training time as illustrated in Figure \ref{fig:graphs}. One thing to note is that we compare CoroNetGAN quantitative results with compression rates of 75$\%$ and 85 $\%$ using CycleGAN architecture unlike $95\%$ in Pix2Pix since it becomes difficult to compress CycleGAN architecture beyond $85\%$ due to its huge model complexity.\par
\textbf{DCGAN:}
We demonstrate the applicability of our approach on unconditional GAN. For the experimentation and evaluating our approach, we adopt DCGAN~\cite{radford2015unsupervised} on CIFAR10 dataset~\cite{krizhevsky2009learning} with original FID score of 45.8. As per the results, we were able to achieve FID score of 56.3 with 50\% compression ratio which outperforms the results obtained with random pruning which achieves FID score of 68.8.
\newline
\textbf{Inference Time Comparisons:}
Additionally, a comparative analysis of the inference time is conducted between the 95\% compressed model obtained from our approach and OMGD~\cite{Ren2021OnlineMD}. Both the models are trained on $\text{Edges}\rightarrow\text{Shoes}$ and evaluated using diverse smartphone chipsets and data types. The inference results, as presented in Table~\ref{tab:multicolts3}, demonstrate that our proposed model achieves superior inference time performance compared to OMGD. 

\subsubsection{Qualitative Results}
We further show visualization results of our proposed method in comparison with state-of-the-art methodologies in Figure~\ref{fig:sample},~\ref{cycle_gan_res} and  \ref{pix2pix_gan_res} demonstrating the effectiveness of our approach. As illustrated, our method can generate high-fidelity images comparable to other state-of-the-art approaches across multiple datasets. The reason we believe our approach generates realistic images is that the compression state of our algorithm forces the generator to generate visually plausible images while competing in the min-max game.
\vspace{-0.3cm}
\subsubsection{Ablation Studies}
Our proposed method for GAN compression shows promising results and outperform state-of-the-art methods on some conditional GANs. We perform extensive ablation studies to further demonstrate the effectiveness of hypernetworks on GAN compression on U-Net based architecture for Pix2Pix. \par
We tried exhaustive hyperparamter search and fine-tuning the learning rate. All the modifications result in a very negligible change of overall FID score.
We also enabled a learning rate scheduler to check the improvement in model performance but quantitatively, no major change was observed. We also tried increasing the layer structure of the hypernetwork employed by increasing the dimension of the embedding space. However, this led to an increase in the training time with a small change in the overall FID score.

\vspace{0.5cm}
\textbf{Compression of both generator and discriminator:}
To evaluate the significance of our approach, we design a variant CoroNetGAN(G+D compression) which compresses both the generator and discriminator till 75$\%$ during training. As mentioned in Table \ref{tab:multicolts2}, this ablation results in improving the FID score from 39.1 to 38.6 as the generator is able to generate better results while compression of discriminator.
\newline
\textbf{Generator finetuning through HyperNetwork:}
We also tried to finetune the weights of the generator generated from the compression state through the HyperNetwork itself, but we did not find any significant improvements in the evaluation metric.


\section{Conclusion}
In this work, we propose a novel method CoroNetGAN for GAN compression based on differentiable pruning via hypernetworks. Unlike other approaches having teacher dependency overhead or post-hoc compression, the compression in our approach is done during the training time itself giving us the benefit of reducing overall time.  Experiments conducted on conditional GANs (Pix2Pix, CycleGAN) substantiate the effectiveness of our proposed method where we have been able to outperform various state-of-the-art techniques on multiple datasets without compromising the visual quality of generated images. Our approach offers significant improvement in training time and inference time as compared to the existing methods. Additionally, we also demonstrate the ability of our approach to be applicable for unconditional GANs (specifically DCGAN). The applicability on other unconditional GANs are open avenues for future work.


{\small
\bibliographystyle{unsrt}
\bibliography{egbib}
}

\end{document}